\pdfoutput=1
\documentclass[11pt,a4paper]{article}
\usepackage{emnlp2018}
\usepackage{times}
\usepackage{latexsym}

\usepackage{url}
\usepackage{graphicx}
\usepackage{subcaption}
\usepackage{natbib}
\usepackage{microtype}
\usepackage{amsmath}
\usepackage{amssymb}

\aclfinalcopy 

\DeclareMathOperator*{\argmax}{arg\,max}

\title{Why are Sequence-to-Sequence Models So Dull?\\[.6ex]
\normalsize Understanding the Low-Diversity Problem of Chatbots}

\author{Shaojie Jiang \\
University of Amsterdam\\
Amsterdam, The Netherlands\\
{\tt s.jiang@uva.nl} \\
\And
Maarten de Rijke \\
University of Amsterdam\\
Amsterdam, The Netherlands\\
{\tt m.derijke@uva.nl} 
}

\date{}

\begin{document}
\maketitle

\begin{abstract}
Diversity is a long-studied topic in information retrieval that usually refers to the requirement that retrieved results should be non-repetitive and cover different aspects. 
In a conversational setting, an additional dimension of diversity matters: an engaging response generation system should be able to output responses that are diverse and interesting.
Sequence-to-sequence (Seq2Seq) models have been shown to be very effective for response generation. 
However, dialogue responses generated by Seq2Seq models tend to have low diversity. 
In this paper, we review known sources and existing approaches to this low-diversity problem. 
We also identify a source of low diversity that has been little studied so far, namely model over-confidence. 
We sketch several directions for tackling model over-confidence and, hence, the low-diversity problem, including confidence penalties and label smoothing.
\end{abstract}

{\let\thefootnote\relax\footnote{{Proceedings of the 2018 EMNLP Workshop SCAI: The 2nd International Workshop on Search-Oriented Conversational AI; ISBN 978-1-948087-75-9}}}


\section{Introduction}

Sequence-to-sequence (Seq2Seq) models \cite{sutskever2014sequence} have been designed for sequence learning.
Generally, a Seq2Seq model consists of two recurrent neural networks (RNN) as its encoder and decoder, respectively, through which the model cannot only deal with inputs and outputs with variable lengths separately, but also be trained end-to-end. 
Seq2Seq models can use different settings for the encoder and decoder networks, such as the number of input/output units, ways of stacking layers, dictionary, etc. 
After showing promising results in machine translation (MT) tasks \cite{sutskever2014sequence, wu2016google}, Seq2Seq models also proved to be effective for tasks like question answering~\cite{yin2015neural}, dialogue response generation~\cite{vinyals2015neural}, text summarization~\cite{nallapati2016abstractive}, constituency parsing~\cite{vinyals2015grammar}, image captioning~\cite{vinyals2015show}, and
so on.

Seq2Seq models form the cornerstone of modern response generation models~\cite{vinyals2015neural, li2015diversity, serban2016building, serban2017hierarchical, zhao2017learning}. 
Although Seq2Seq models can generate grammatical and fluent responses, it has also been reported that the corpus-level diversity of Seq2Seq models is usually low, as many responses are trivial or non-committal, like ``I don't know'', ``I'm sorry'' or ``I'm OK''~\cite{vinyals2015neural, sordoni2015neural, serban2016building, li2015diversity}. 
We refer to this problem as the \emph{low-diversity} problem.

In recent years, there have been several types of approach to diagnosing and addressing the low-diversity problem. 
The purpose of this paper is to understand the low-diversity problem, to understand what diagnoses and solutions have been proposed so far, and to explore possible new approaches. 
We first review the theory of Seq2Seq models, then we give an overview of known
causes and existing solutions to the low-diversity problem. 
We then connect the low-diversity problem to the concept of \emph{model over-confidence}, and propose approaches to address the over-confidence problem and, hence, the low-diversity problem. 



\section{Sequence-to-Sequence Response Generation}

Consider a dataset of message-response pairs $(X, Y)$, where $X=(x_1, x_2, \dots, x_{|X|})$ and $Y=(y_1, y_2, \dots, y_{|Y|})$ are the input and output sequences, respectively. 
During training, the goal is to learn the relationships between $X$ and $Y$, which can be formulated as maximizing the Seq2Seq model probability of $Y$ given $X$:
\begin{equation}
  \label{eq:tok}
  \max p(Y|X) = \max \prod_{t=1}^{|Y|} p(y_t| y_{<t}, X),
\end{equation}
where $y_{<t} = (y_1, y_2, \dots, y_{t-1})$ are the ground-truth tokens of previous steps.

Usually, Seq2Seq models employ Long Short-Term Memory (LSTM) networks as their encoder and decoder. 
The way a Seq2Seq models realizes \eqref{eq:tok}, is to process the training inputs and outputs separately. 
On the encoder side, the input sequence $X$ is encoded step-by-step, e.g., at step $t$:
\begin{equation}
  \label{eq:encoder}
  h_t^{enc} = f_\theta^{enc}(h_{t-1}^{enc}, x_t),
\end{equation}
where $h_0^{enc} = \mathbf{0}$ is the initial hidden state of the encoder LSTM, and $\theta$ is the model parameter. 
The hidden state of the last step $h_{|X|}^{enc}$ is the vector representation of input sequence $X$.

Then, the decoder LSTM is initialized by $h_0^{dec} = h_{|X|}^{enc}$ so that output tokens can be based on the input:
\begin{equation}
  \label{eq:decoder}
  h_t^{dec} = f_\theta^{dec}(h_{t-1}^{dec}, y_{t-1}),
\end{equation}
with $y_0$ as a special token (e.g., \_START\_) to indicate the decoder to start generation, and $y_{t-1}$ as the ground truth token of the last time step. 
The hidden state $h_t^{dec}$ is further used to predict the output distribution by using a multi-layer perceptron (MLP) and softmax function:
\begin{equation}
  \label{eq:prob}
  \mbox{}\hspace*{-1mm}
  P(y_t|y_{<t}, X) =
  \frac{\exp(c_i f_\theta^{MLP}(h_t^{dec}))}{\sum_{j=1}^N
    \exp(c_j f_\theta^{MLP}(h_t^{dec}))},
\end{equation}
where $c_*$ are possible candidates of $y_t$, which are usually represented as word embeddings. 
After obtaining this distribution, we can calculate the loss compared with the ground-truth distribution by using, e.g., the cross-entropy loss function, and then we can back-propagate the loss to force the Seq2Seq model to maximize \eqref{eq:tok}.

At test time at $t$, the step-wise decoder output distribution is conditioned on the actual model outputs $\hat{y}_{<t}$ and $X$, and the token with the highest probability is chosen as the output:
\begin{equation}
  \hat{y}_t = \argmax_{y_t} p(y_t|\hat{y}_{<t},X),
\end{equation}
which is known as the maximum \emph{a posteriori} (MAP) objective function.



\section{Diagnosing the Low-Diversity Problem}

In the literature, three dominant viewpoints on the low-diversity problem have been shared: lack of variability, improper objective function, and weak conditional signal. 
Below, we review these diagnoses of the low-diversity problem, with corresponding solutions, and we add a fourth diagnosis: model over-confidence.

\subsection{Lack of variability}

\citet{serban2017hierarchical, zhao2017learning} trace the cause of the low-diversity problem in Seq2Seq models back to the lack of model variability. The variability of Seq2Seq models is different from that of retrieval-based chatbots \cite{fedorenko2017avoiding}: in this study, we focus on the lack of variability of system responses, while in \cite{fedorenko2017avoiding}, the authors deal with the low variability between responses and contexts.

To increase variability, \citet{serban2017hierarchical, zhao2017learning} propose to introduce variational autoencoders (VAEs) to Seq2Seq models. 
At generation time, the latent variable $z$ brought by a VAE is used as a conditional signal of the decoder LSTM~\cite{serban2017hierarchical}:
\begin{equation}
  \label{eq:latent}
  h_t^{dec} = f_\theta^{dec}(h_{t-1}^{dec}, y_{t-1}, z),
\end{equation}
where we omit the contextual hidden states for simplicity.

At test time, $z$ is \emph{randomly} sampled from a prior distribution. 
Although being effective, the improvement in the degree of diversity of generated responses brought by this kind of method is actually brought by the randomness of $z$. 
The underlying Seq2Seq model remains sub-optimal in terms of diversity.

\subsection{Improper objective function}

\citet{li2015diversity} notice that the MAP objective function may be the cause of the low-diversity problem, since it can favor certain responses by only maximizing $p(Y|X)$. 
Therefore, they propose to maximize the mutual information between $X$, $Y$ pairs:
\begin{equation}
  \label{eq:MMI}
  \log \frac{p(X, Y)}{p(X)p(Y)}.
\end{equation}
With the help of Bayes' theorem, they derive two Maximum Mutual Information (MMI) objective functions:
\begin{equation}
  \label{eq:MMI-antiLM}
\begin{split}  
  \hat{Y} =\argmax_{Y} \{&\log p(Y|X) -{}\\
  & \lambda \log p(Y) + \gamma |Y|\},
\end{split}
\end{equation}
and
\begin{equation}
  \label{eq:MMI-bidi}
  \begin{split}
    \hat{Y} = \argmax_{Y} \{&(1-\lambda)\log p(Y|X) + \\
    &\lambda \log p(X|Y) + \gamma |Y|\},
  \end{split}
\end{equation}
where $\lambda$ and $\gamma$ are hyper-parameters. 
Here, $\log p(Y)$ and $\log p(X|Y)$ are the language model and a reverse model, respectively, with the latter trained using response-message pairs: $(Y, X)$. 
Besides the time needed for training a reverse model, it should be noted that both objective functions need the length $|Y|$ of candidate responses, which are maintained in N-best lists generated by beam search. 
To obtain N-best lists with enough diversity, \citet{li2015diversity} use a beam size of 200 during testing, which is much more time-consuming than the basic Seq2Seq model.

Influenced by the MMI methods, several beam search based approaches \cite{li2016simple, vijayakumar2016diverse, shao2017generating} focus on improving the diversity of N-best lists, in the hope of further enhancing the one-best response diversity. 
However, there are other faster approaches to the low-diversity problem without using beam
search, such as the attention-based model that we describe below.

\subsection{Weak conditional signal}

Since attention layers \cite{bahdanau2014neural} have been introduced into Seq2Seq models for the MT task, they have also been a \emph{de facto} standard module of Seq2Seq models for response generation. 
The purpose of Seq2Seq attention layers is different from the purpose of the Transformer model \cite{vaswani2017attention}. 
Transformer proposes to rely only on self-attention and avoid using reccurence or convolutions, while attention layers of Seq2Seq aim at strengthening the input signal.

Although the introducing of attention layers can bring improvements to the response generation task, \citet{tao2018get} argue that the original attention signal often focuses on particular parts of the input sequence, which is not strong enough for the Seq2Seq model to generate specific responses, thus causing the low-diversity problem. 
The authors propose to use multiple attention heads to encourage the model to focus on various aspects of the input, by mapping encoder hidden states to $K$ different semantic spaces:
\begin{equation}
  \label{eq:MHAM}
  h_{t, k}^{enc} = W_p^k \cdot h_t^{enc},
\end{equation}
where $W_p^k \in \mathbb{R}^{d \times d}$ is a learnable projection matrix. 
The net effect of the extended attention mechanism is, indeed, improvements in the diversity of generated responses.
Readers are referred to \cite{tao2018get} for more details.



\subsection{Model over-confidence}

As indicated by \citet{hinton2015distilling}, one can think of the knowledge captured in conversation modeling as a mapping from input sequence $X$ to output sequence $Y$, i.e., the distribution $P(Y|X)$. 
Therefore, if responses have a low degree of diversity, the learned distribution $P(Y|X)$ is questionable, as re-confirmed by~\citet{li2015diversity}.
According to \eqref{eq:tok}, the sequence-level distribution $P(Y|X)$ has a direct relationship with the token-level distribution. 
Therefore, we hypothesize that the token-level distribution $P(y_t | y_{<t}, X)$, produced at the decoder side, may be the culprit.

The decoder LSTM serves as an RNN language model (RNNLM) conditioned on the input sequence \cite{sutskever2014sequence}. 
With time steps increasing, the influence of the input sequence $X$ will become weaker according to \eqref{eq:decoder}, and if the token-level distribution $P(y_t| y_{<t}, X)$ is problematic, it will have further effects on subsequent outputs (a ``snowball effect''). 
An attention mechanism \cite{bahdanau2014neural, tao2018get} can be used to reinforce the influence of the input sequence, but there are still chances that the detrimental effect of $P(y_t| y_{<t}, X)$ is stronger than the input signal.

\begin{figure}[ht!]
  \centering
  \includegraphics[width=\columnwidth]{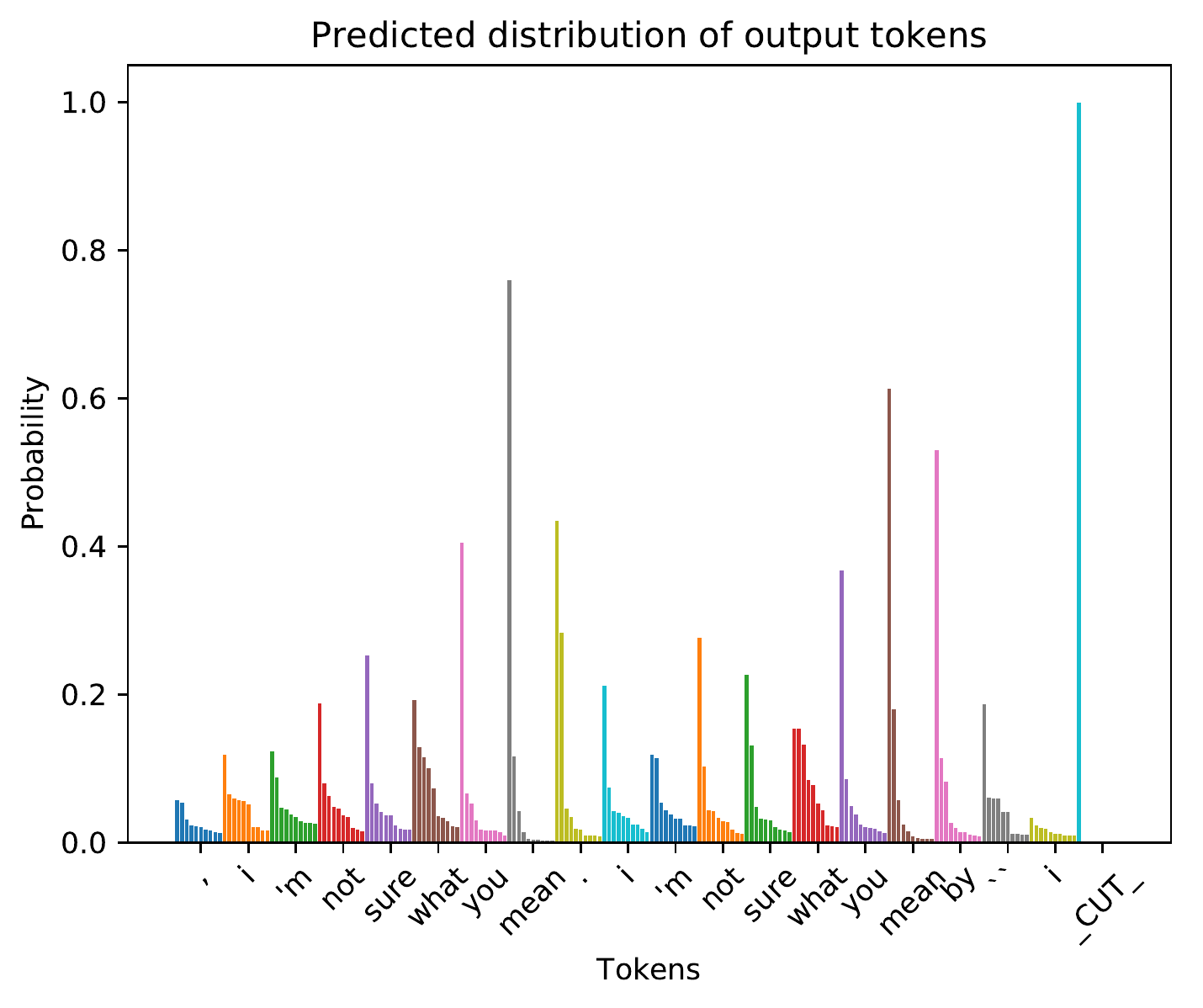}
  \caption{Given the input sequence: \emph{how about we recognize the
      brilliance in everyone, or in mankind as a whole.}, the
    predicted distribution of model outputs, and tokens on x axis are
    MAP predictions. Note that we kept top-10 probabilities at each
    prediction step for simplicity and this output was cut before the
    \_EOS\_ token was emitted.}
  \label{fig:noattn}
\end{figure}

\begin{figure}[ht!]
  \centering
  \includegraphics[width=\columnwidth]{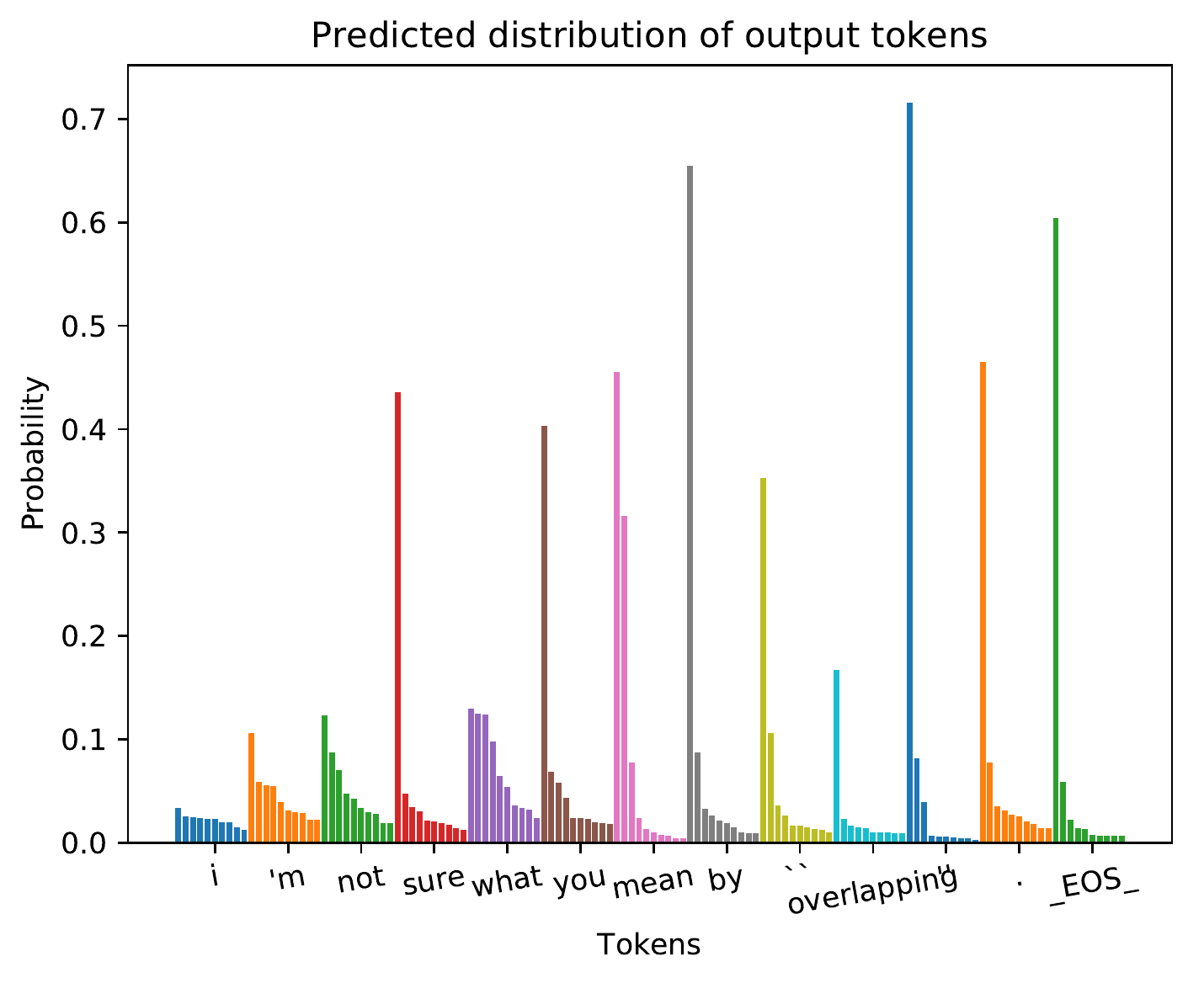}
  \caption{Predicted distribution of the same input as in Figure
    \ref{fig:noattn} when an attention mechanism is used.}
  \label{fig:attn}
\end{figure}

To analyze the problem of $P(y_t| y_{<t}, X)$, we train a Seq2Seq model\footnote{We are using ParlAI framework~\cite{miller2017parlai}.} without attention layer, and plot the
token-level distribution of generic responses in Figure~\ref{fig:noattn}. 
Interestingly, we find that the distributions shown signs of model over-confidence
\cite{pereyra2017regularizing}. 
When an attention mechanism is used, similar distributions can still be observed, as illustrated in
Figure \ref{fig:attn}. 
From these two figures, we can see a common trend of growing confidence: the highest probabilities at each step keep growing, which confirms our conjecture of a snowball effect. 
Due to this effect, the final several tokens are of low quality, e.g., the no-attention model in Figure~\ref{fig:noattn} starts to repeat itself, and the word ``overlapping'' in the attention model in Figure~\ref{fig:attn} is irrelevant for the user input.

A prediction is confident if the entropy of the output distribution is low.
\emph{Over-confidence} is often a symptom of over-fitting \cite{szegedy2016rethinking}, which suggests that the inputs or outputs share much similarity from unknown aspects. 
Although it is hard to figure out what causes the over-fitting, maximizing entropy can usually help to regularize the model, making it generalize better. 
In~\cite{pereyra2017regularizing}, the authors propose to add the negative entropy to the negative log-likelihood loss function during training, which can easily be tailored for conversation modeling:
\begin{equation}
  \label{eq:conf_pen}
  \begin{split}
    \mathcal{L} (\theta) = - \sum_{i=1}^N &\log p(c_i|y_{<t}, X) - \\
    &\beta H(p(c_i|y_{<t}, X)),
  \end{split}
\end{equation}
where $\beta$ controls the strength of the confidence penalty, and $H(\cdot)$ is the entropy of the output distribution:
\begin{equation}
  \label{eq:entropy}
  \begin{aligned}
    &H(p(c_i|y_{<t}, X)) = \\
    &\mbox{}\quad- \sum_{i=1}^N p(c_i|y_{<t}, X) \log (p(c_i|y_{<t}, X)).
  \end{aligned}
\end{equation}
The authors also show that this confidence penalty method is closely related to label smoothing regularization \cite{szegedy2016rethinking}, therefore methods like neighborhood smoothing \cite{chorowski2016towards} may be used to solve the low-diversity problem.

So far, there has been no published work on analyzing the the effectiveness of correcting for model over-confidence on the low-diversity problem.
It is important to note the fourth diagnosis of the low-diversity problem, i.e., that the problem is due to model over-confidence, is essentially different from the three types of diagnosis that we described earlier in the section.
Among diagnoses and methods published previously, the VAE-based approaches actually bypass the low-diversity problem by introducing randomness; MMI-based methods
have an elegant theoretical basis, yet they end up relying on many extra modules, like reverse models and beam search, and the newly-introduced hyper-parameters were not even learned from training data \cite{li2015diversity}; attention-based models offer a complementary approach, since strengthening the conditional signal is likely to make the response more specific, which should in turn improve the corpus-level diversity. 
Model over-confidence may offer a simpler alternative -- we believe that methods such as confidence penalty are likely to alleviate the low-diversity problem in ways that differ from previous approaches.



\section{Next Steps}

In this paper, we described the low-diversity problem for response generation, which is one of
the main issues faced by current Seq2Seq-based conversation models. 
We reviewed existing diagnoses and corresponding approaches to this problem and also added a diagnosis that has not been proposed or used so far, i.e., model over-confidence.

By using entropy maximizing approaches, such as confidence penalty \cite{pereyra2017regularizing} or label smoothing \cite{szegedy2016rethinking}, we believe that the low-diversity problem of Seq2Seq models can be alleviated. 
Besides, by using entropy maximizing methods, the self-repeating problem \cite{li2017adversarial} of Seq2Seq models may also be alleviated since this can reduce the snowball effect and make later outputs more relevant. 
We also noticed that the low-diversity problem resembles the mode collapse problem of GANs \cite{goodfellow2014generative}, therefore inspirations may be drawn from the solutions like
\cite{salimans2016improved, metz2016unrolled}.

In addition, since we now have four types of diagnosis of the low-diversity problem, each of which is likely to address part of the problem but not all of the problem, it is natural to systematically compare and combine approaches based on the different types of diagnosis. 
Understanding how solutions to the low-diversity problem helps to improve the effectiveness of conversational agents for search-oriented tasks is another interesting line of future work.


\section*{Acknowledgments}
This research was supported by the China Scholarship Council.

\bibliography{seq2seq}

\begin{thebibliography}{28}
\expandafter\ifx\csname natexlab\endcsname\relax\def\natexlab#1{#1}\fi

\bibitem[{Bahdanau et~al.(2014)Bahdanau, Cho, and Bengio}]{bahdanau2014neural}
Dzmitry Bahdanau, Kyunghyun Cho, and Yoshua Bengio. 2014.
\newblock Neural machine translation by jointly learning to align and
  translate.
\newblock \emph{arXiv preprint arXiv:1409.0473}.

\bibitem[{Chorowski and Jaitly(2016)}]{chorowski2016towards}
Jan Chorowski and Navdeep Jaitly. 2016.
\newblock Towards better decoding and language model integration in sequence to
  sequence models.
\newblock \emph{arXiv preprint arXiv:1612.02695}.

\bibitem[{Fedorenko et~al.(2017)Fedorenko, Smetanin, and
  Rodichev}]{fedorenko2017avoiding}
Denis Fedorenko, Nikita Smetanin, and Artem Rodichev. 2017.
\newblock Avoiding echo-responses in a retrieval-based conversation system.
\newblock \emph{arXiv preprint arXiv:1712.05626}.

\bibitem[{Goodfellow et~al.(2014)Goodfellow, Pouget-Abadie, Mirza, Xu,
  Warde-Farley, Ozair, Courville, and Bengio}]{goodfellow2014generative}
Ian Goodfellow, Jean Pouget-Abadie, Mehdi Mirza, Bing Xu, David Warde-Farley,
  Sherjil Ozair, Aaron Courville, and Yoshua Bengio. 2014.
\newblock Generative adversarial nets.
\newblock In \emph{Advances in neural information processing systems}, pages
  2672--2680.

\bibitem[{Hinton et~al.(2015)Hinton, Vinyals, and Dean}]{hinton2015distilling}
Geoffrey Hinton, Oriol Vinyals, and Jeff Dean. 2015.
\newblock Distilling the knowledge in a neural network.
\newblock \emph{arXiv preprint arXiv:1503.02531}.

\bibitem[{Li et~al.(2015)Li, Galley, Brockett, Gao, and
  Dolan}]{li2015diversity}
Jiwei Li, Michel Galley, Chris Brockett, Jianfeng Gao, and Bill Dolan. 2015.
\newblock A diversity-promoting objective function for neural conversation
  models.
\newblock \emph{arXiv preprint arXiv:1510.03055}.

\bibitem[{Li et~al.(2016)Li, Monroe, and Jurafsky}]{li2016simple}
Jiwei Li, Will Monroe, and Dan Jurafsky. 2016.
\newblock A simple, fast diverse decoding algorithm for neural generation.
\newblock \emph{arXiv preprint arXiv:1611.08562}.

\bibitem[{Li et~al.(2017)Li, Monroe, Shi, Ritter, and
  Jurafsky}]{li2017adversarial}
Jiwei Li, Will Monroe, Tianlin Shi, Alan Ritter, and Dan Jurafsky. 2017.
\newblock Adversarial learning for neural dialogue generation.
\newblock \emph{arXiv preprint arXiv:1701.06547}.

\bibitem[{Metz et~al.(2016)Metz, Poole, Pfau, and
  Sohl-Dickstein}]{metz2016unrolled}
Luke Metz, Ben Poole, David Pfau, and Jascha Sohl-Dickstein. 2016.
\newblock Unrolled generative adversarial networks.
\newblock \emph{arXiv preprint arXiv:1611.02163}.

\bibitem[{Miller et~al.(2017)Miller, Feng, Fisch, Lu, Batra, Bordes, Parikh,
  and Weston}]{miller2017parlai}
Alexander~H Miller, Will Feng, Adam Fisch, Jiasen Lu, Dhruv Batra, Antoine
  Bordes, Devi Parikh, and Jason Weston. 2017.
\newblock Parlai: A dialog research software platform.
\newblock \emph{arXiv preprint arXiv:1705.06476}.

\bibitem[{Nallapati et~al.(2016)Nallapati, Zhou, Gulcehre, Xiang
  et~al.}]{nallapati2016abstractive}
Ramesh Nallapati, Bowen Zhou, Caglar Gulcehre, Bing Xiang, et~al. 2016.
\newblock Abstractive text summarization using sequence-to-sequence rnns and
  beyond.
\newblock \emph{arXiv preprint arXiv:1602.06023}.

\bibitem[{Pereyra et~al.(2017)Pereyra, Tucker, Chorowski, Kaiser, and
  Hinton}]{pereyra2017regularizing}
Gabriel Pereyra, George Tucker, Jan Chorowski, {\L}ukasz Kaiser, and Geoffrey
  Hinton. 2017.
\newblock Regularizing neural networks by penalizing confident output
  distributions.
\newblock \emph{arXiv preprint arXiv:1701.06548}.

\bibitem[{Salimans et~al.(2016)Salimans, Goodfellow, Zaremba, Cheung, Radford,
  and Chen}]{salimans2016improved}
Tim Salimans, Ian Goodfellow, Wojciech Zaremba, Vicki Cheung, Alec Radford, and
  Xi~Chen. 2016.
\newblock Improved techniques for training gans.
\newblock In \emph{Advances in Neural Information Processing Systems}, pages
  2234--2242.

\bibitem[{Serban et~al.(2016)Serban, Sordoni, Bengio, Courville, and
  Pineau}]{serban2016building}
Iulian~Vlad Serban, Alessandro Sordoni, Yoshua Bengio, Aaron~C Courville, and
  Joelle Pineau. 2016.
\newblock Building end-to-end dialogue systems using generative hierarchical
  neural network models.
\newblock In \emph{AAAI}, volume~16, pages 3776--3784.

\bibitem[{Serban et~al.(2017)Serban, Sordoni, Lowe, Charlin, Pineau, Courville,
  and Bengio}]{serban2017hierarchical}
Iulian~Vlad Serban, Alessandro Sordoni, Ryan Lowe, Laurent Charlin, Joelle
  Pineau, Aaron~C Courville, and Yoshua Bengio. 2017.
\newblock A hierarchical latent variable encoder-decoder model for generating
  dialogues.
\newblock In \emph{AAAI}, pages 3295--3301.

\bibitem[{Shao et~al.(2017)Shao, Gouws, Britz, Goldie, Strope, and
  Kurzweil}]{shao2017generating}
Louis Shao, Stephan Gouws, Denny Britz, Anna Goldie, Brian Strope, and Ray
  Kurzweil. 2017.
\newblock Generating high-quality and informative conversation responses with
  sequence-to-sequence models.
\newblock \emph{arXiv preprint arXiv:1701.03185}.

\bibitem[{Sordoni et~al.(2015)Sordoni, Galley, Auli, Brockett, Ji, Mitchell,
  Nie, Gao, and Dolan}]{sordoni2015neural}
Alessandro Sordoni, Michel Galley, Michael Auli, Chris Brockett, Yangfeng Ji,
  Margaret Mitchell, Jian-Yun Nie, Jianfeng Gao, and Bill Dolan. 2015.
\newblock A neural network approach to context-sensitive generation of
  conversational responses.
\newblock \emph{arXiv preprint arXiv:1506.06714}.

\bibitem[{Sutskever et~al.(2014)Sutskever, Vinyals, and
  Le}]{sutskever2014sequence}
Ilya Sutskever, Oriol Vinyals, and Quoc~V Le. 2014.
\newblock Sequence to sequence learning with neural networks.
\newblock In \emph{Advances in neural information processing systems}, pages
  3104--3112.

\bibitem[{Szegedy et~al.(2016)Szegedy, Vanhoucke, Ioffe, Shlens, and
  Wojna}]{szegedy2016rethinking}
Christian Szegedy, Vincent Vanhoucke, Sergey Ioffe, Jon Shlens, and Zbigniew
  Wojna. 2016.
\newblock Rethinking the inception architecture for computer vision.
\newblock In \emph{Proceedings of the IEEE conference on computer vision and
  pattern recognition}, pages 2818--2826.

\bibitem[{Tao et~al.(2018)Tao, Gao, Shang, Wu, Zhao, and Yan}]{tao2018get}
Chongyang Tao, Shen Gao, Mingyue Shang, Wei Wu, Dongyan Zhao, and Rui Yan.
  2018.
\newblock Get the point of my utterance! learning towards effective responses
  with multi-head attention mechanism.
\newblock In \emph{IJCAI}, pages 4418--4424.

\bibitem[{Vaswani et~al.(2017)Vaswani, Shazeer, Parmar, Uszkoreit, Jones,
  Gomez, Kaiser, and Polosukhin}]{vaswani2017attention}
Ashish Vaswani, Noam Shazeer, Niki Parmar, Jakob Uszkoreit, Llion Jones,
  Aidan~N Gomez, {\L}ukasz Kaiser, and Illia Polosukhin. 2017.
\newblock Attention is all you need.
\newblock In \emph{Advances in Neural Information Processing Systems}, pages
  6000--6010.

\bibitem[{Vijayakumar et~al.(2016)Vijayakumar, Cogswell, Selvaraju, Sun, Lee,
  Crandall, and Batra}]{vijayakumar2016diverse}
Ashwin~K Vijayakumar, Michael Cogswell, Ramprasath~R Selvaraju, Qing Sun,
  Stefan Lee, David Crandall, and Dhruv Batra. 2016.
\newblock Diverse beam search: Decoding diverse solutions from neural sequence
  models.
\newblock \emph{arXiv preprint arXiv:1610.02424}.

\bibitem[{Vinyals et~al.(2015{\natexlab{a}})Vinyals, Kaiser, Koo, Petrov,
  Sutskever, and Hinton}]{vinyals2015grammar}
Oriol Vinyals, {\L}ukasz Kaiser, Terry Koo, Slav Petrov, Ilya Sutskever, and
  Geoffrey Hinton. 2015{\natexlab{a}}.
\newblock Grammar as a foreign language.
\newblock In \emph{Advances in Neural Information Processing Systems}, pages
  2773--2781.

\bibitem[{Vinyals and Le(2015)}]{vinyals2015neural}
Oriol Vinyals and Quoc Le. 2015.
\newblock A neural conversational model.
\newblock \emph{arXiv preprint arXiv:1506.05869}.

\bibitem[{Vinyals et~al.(2015{\natexlab{b}})Vinyals, Toshev, Bengio, and
  Erhan}]{vinyals2015show}
Oriol Vinyals, Alexander Toshev, Samy Bengio, and Dumitru Erhan.
  2015{\natexlab{b}}.
\newblock Show and tell: A neural image caption generator.
\newblock In \emph{Proceedings of the IEEE conference on computer vision and
  pattern recognition}, pages 3156--3164.

\bibitem[{Wu et~al.(2016)Wu, Schuster, Chen, Le, Norouzi, Macherey, Krikun,
  Cao, Gao, Macherey et~al.}]{wu2016google}
Yonghui Wu, Mike Schuster, Zhifeng Chen, Quoc~V Le, Mohammad Norouzi, Wolfgang
  Macherey, Maxim Krikun, Yuan Cao, Qin Gao, Klaus Macherey, et~al. 2016.
\newblock Google's neural machine translation system: Bridging the gap between
  human and machine translation.
\newblock \emph{arXiv preprint arXiv:1609.08144}.

\bibitem[{Yin et~al.(2015)Yin, Jiang, Lu, Shang, Li, and Li}]{yin2015neural}
Jun Yin, Xin Jiang, Zhengdong Lu, Lifeng Shang, Hang Li, and Xiaoming Li. 2015.
\newblock Neural generative question answering.
\newblock \emph{arXiv preprint arXiv:1512.01337}.

\bibitem[{Zhao et~al.(2017)Zhao, Zhao, and Eskenazi}]{zhao2017learning}
Tiancheng Zhao, Ran Zhao, and Maxine Eskenazi. 2017.
\newblock Learning discourse-level diversity for neural dialog models using
  conditional variational autoencoders.
\newblock \emph{arXiv preprint arXiv:1703.10960}.

\end{thebibliography}
\bibliographystyle{acl_natbib_nourl}

\end{document}